# Machine Learning Detection Algorithm for Large Barkhausen Jumps in Cluttered Environment

Roger Alimi[1], Amir Ivry[2], Elad Fisher[1,3], and Eyal Weiss[1]

[1] *Technology Division, Soreq NRC, Yavne 81800, Israel*
[2] *Technion, Israel Institute of Technology, Haifa 32000, Israel*
[3] *Jerusalem College of Technology, Jerusalem 91160, Israel*

*Abstract*—Modern magnetic sensor arrays conventionally utilize state of the art low power magnetometers such as parallel and orthogonal fluxgates. Low power fluxgates tend to have large Barkhausen jumps that appear as a dc jump in the fluxgate output. This phenomenon deteriorates the signal fidelity and effectively increases the internal sensor noise. Even if sensors that are more prone to dc jumps can be screened during production, the conventional noise measurement does not always catch the dc jump because of its sparsity. Moreover, dc jumps persist in almost all the sensor cores although at a slower but still intolerable rate. Even if dc jumps can be easily detected in a shielded environment, when deployed in presence of natural noise and clutter, it can be hard to positively detect them. This work fills this gap and presents algorithms that distinguish dc jumps embedded in natural magnetic field data. To improve robustness to noise, we developed two machine learning algorithms that employ temporal and statistical physical-based features of a pre-acquired and well-known experimental data set. The first algorithm employs a support vector machine classifier, while the second is based on a neural network architecture. We compare these new approaches to a more classical kernel-based method. To that purpose, the receiver operating characteristic curve is generated, which allows diagnosis ability of the different classifiers by comparing their performances across various operation points. The accuracy of the machine learning-based algorithms over the classic method is highly emphasized. In addition, high generalization and robustness of the neural network can be concluded, based on the rapid convergence of the corresponding receiver operating characteristic curves.

## I. INTRODUCTION

Fluxgate magnetometers are induction sensors employing a soft magnetic core which is periodically saturated [Ripka 2001]. There are two main types of fluxgates; A parallel fluxgate [Janosek 2017] where the core is excited by magnetic field parallel to the measured field, and an orthogonal fluxgate [Butta 2017] where it is orthogonal. Parallel fluxgates cores are excited by a bipolar magnetic field where orthogonal fluxgates [Primdahl 1979] are modulated by a unipolar field [Paperno 2004].

When employing a sensor array it is important to maintain low power consumption. However, it is known that the internal noise increases when decreasing the fluxgate excitation field [Musmann 2010]. As a result, the fluxgate core does not undergo deep and uniform saturation [Weiss 2014] and the magnetization of the core is inhomogeneous. As a result, low power fluxgates tend to suffer from dc jumps in their output.

A physical description of jumps in the output of low power fluxgates is first introduced by the authors in [Weiss 2018]. We have shown that because of the low saturation excitation field, some magnetic domains are temporarily "stuck" in one magnetization direction. They are stuck on metallurgical imperfections in the lattice [Weiss 2014], while the rest of the core domains continue in periodical rotations. The "stuck" domains disturb the effective permeability of the core, which is translated to a rapid change in fluxgate sensitivity. An experimentally physical model for the dc jumps based on an expansion of the Landau Lifshitz Gilbert (LLG) equation and an expansion to include the core excitation dynamics has been already presented [Weiss 2019]. Regardless of the physical origin of dc jumps, from a practical, applicable point of view [Alimi 2009], we believe it is crucial to be able not only to detect, but also to distinguish dc jumps from very similar signal patterns. These patterns have to be treated based on the specific application and cannot be confused with the Barkhausen phenomenon.

A dc jump in the fluxgate output deteriorates the signal fidelity and effectively increases the internal sensor noise. The dc jump is a step-like phenomenon with a sparse and stochastic pattern, as illustrated in Fig. 1. Sensors that are more prone to dc jumps can be screened during production by performing internal noise measurement in a magnetically shielded chamber and employing either an entropy detector or kernel-based methods to detect the dc jumps. However, the conventional noise measurement does not always catch the dc jump because of its sparsity. Nevertheless, dc jumps persist in almost all the sensor cores although at a slower rate [Weiss 2019]. As a result, low-power consuming fluxgates cannot be utilized to their full potential because their output is afflicted with dc jumps that severely impede their performance.

In applications like surveillance systems DC jumps can compromise both detection and localization of relevant signals. They are well described in [Alimi 2009] and [Alimi 2015], see also the work of Kozick and Sadler in [Kozick 2008]. Another field of application is the magnetic localization of wireless capsule endoscopy where unwanted DC changes can induce large localization errors. See for instance the work of Pham and Aziz in [Pham 2014].

Although dc jumps are easily detected in a shielded environment, when deployed in an environment of natural noise and clutter, they are difficult to be positively detected. In this work we bridge this gap and present algorithms that distinguish dc jumps embedded in natural magnetic field measurements. This is important because it improves the sensors' signal fidelity and enables a more precise and performant characterization of the signal.

The paper is organized as follows. Section II describes three detection algorithms we have developed. Section III presents an experimental setting from which the database was created, the setup design and the training process of the algorithms. In Section IV we discuss the results, and Section V concludes.

## II. DETECTION ALGORITHMS

In this study, we draw comparisons between three algorithms for dc jumps detection. These algorithms can be divided into two groups; a classic kernel-based method and learning-based algorithms.

### A. Kernel-based

This classic approach is based on the low ordered statistical nature of the signal and is inspired by Canny's edge detector [Canny 1986], now reduced to one dimension. Here, a template matching process is employed in the time domain between a dedicated kernel and the magnetic measurements. This template then undergoes statistical analysis, in which anomalies are detected. In this study, we employ a detection kernel $K(n, \sigma)$, which is the derivative of a Gaussian with variance $\sigma^2$ and zero mean. Tweaking the parameter $\sigma$ controls the sharpness of jumps this method can detect. To better grasp the core of this method, let us lay out the following intuitive concept; if $\sigma$ is very small, only extremely sharp changes in the signal are laid out from the rest of the signal. On the other hand, jumps spread across relatively large number of samples may bear high resemblance to other parts of the template, which calls for large values of $\sigma$ to obtain high performance.

### B. Support Vector Machine

Since the kernel-based method relies on empirical parametric optimization, it suffers from high sensitivity to noise. The proposed learning-based support vector machine method (SVM) [Cortes 1995] deals with this issue by generating a classifier that exploits nine temporal and physical features of the measured magnetic signals.

The research of magnetic signals in this study revealed that linear operations cannot transfer one type of signal (e.g. jump) to the second type of signal. Namely, there are non-linear relations between the two. Thus, in order to distinguish between them, one must derive non-linear relations from the measurements, and establish the classifier on them.

### C. Artificial Neural Network

Even though SVM can produce generalizing and robust models to a degree, its ability to handle highly complex relations is limited by its single non-linearity modelling (parabolic, radial, etc.). This drawback leads to unsatisfactory robustness and calls for deep learning solutions. In this study, we design a three-layered artificial neural network (ANN), with 12 neurons in each hidden layer [Ciresan 2012]. The features used for the SVM remain unchanged and they are inserted into an input layer comprise of 9 neurons. The output layer produces a 1-bit indicator for the presence of jumps. The activation function in the end of each neuron is the Rectifier Linear Unit (ReLU) [Nair 2010] function.

On the other hand, the strength of the ANN can be its Achilles hill, and causes overfitting. To overcome that, we both add a regularization term to the objective function and employ the dropout technique [Srivastava 2014]. The former limits the values of the parameters in the network and maintains them to a low dynamic range. The latter effectively reduces the number of neurons used in the training process of the network, so the model can be parametrized with fewer coefficients. Formally, let the training set be the matrix $F^{tr} \in \mathbb{R}^{m \times 10}$. It comprises $m$ training measurements, where each contains 9 features and a 1-bit label that indicates whether this measurement is truly a jump or not. Also, let us notate the non-linear output of the network as $Z_{net}(f_i^{tr}) \in \{0,1\}$, where $f_i^{tr} = [f_{i1}^{tr}, \dots, f_{i9}^{tr}]$ is the $i^{th}$ input feature vector to the ANN and $1 \leq i \leq m$. Therefore, the objective function $J(F, w)$ can be defined as:

$$J(F, w) = \sum_{i=1}^{m} \left\| Z_{net}(f_i^{tr}) - f_{i,10} \right\|^2 + \lambda \sum_{n=1}^{N} w_n^2 ,$$

where $N$ is the number of parameters in the ANN, $w = [w_1, \dots, w_N]$ is the vector that contains their values, and $\lambda$ is a strictly positive number that controls the weight of the regularization term. Respectively, let us establish the following optimization problem to be solved:

$$C_{net} = \underset{w \in \mathbb{R}^N}{\mathrm{argmin}}\, J(F, w) .$$

So, the following mapping is applied by the trained network on a given feature vector $f = [f_1, \dots, f_9]$:

$$C_{net}(f) = \begin{cases} jump, & if\ Z_{net}(f) = 1 \\ not\ jump, & if\ Z_{net}(f) = 0 \end{cases}.$$

## III. EXPERIMENTAL SETTINGS

### A. Apparatus

Data was acquired from an array of $24$ Bartington's Mag648 magnetometers [Mag648 2011], of parallel 3-axial fluxgate operating with a bipolar excitation. The sensor sensitivity is $50\ mV/\mu T$ and the typical internal noise density is smaller than $20\ pT/\sqrt{Hz}$ at $1\ Hz$. They were digitized by a dedicated $24$ bit data acquisition and sampled at a rate of $10\ Hz$.

### B. Database and Features

We divide the database into three different categories; training set, evaluation set and test set, containing $600$, $100$ and $100$ magnetic measurements correspondingly. Every measurement includes $450$ samples, and all sets are balanced, in the sense that they comprise the same number of jumps and non-jumps sequences. Also, for each sequence of the 800, we attach a 1-bit label of 0 or 1, which indicates the absence or presence of a jump in that sequence, respectively.

Formally, let us notate the set of sequences as $\{s_i\}_{i=1}^{M} \in \mathbb{R}^{450}$ and their corresponding label set as $\{\ell_i\}_{i=1}^{M} \in \{0,1\}$, where $M = 800$. For each measurement $s_i$, we perform a feature extraction process, and obtain a 9-dimensional vector $\boldsymbol{f}_i \in \mathbb{R}^9$. Now, the set $\{\boldsymbol{f}_i, \ell_i\}_{i=1}^{M} \in \mathbb{R}^{10}$ is the sole information to be carried out and inserted into the different algorithms. This database goes through a feature extraction process that includes two types of features; statistical features and temporal features, where the latter is always casual in order to avoid system delay. The features are described in detail in Table 1, and they are numbered in descending order according to their contribution to the test accuracy. Before concluding this section, an important remark is to be made.

Although the output of this vector fluxgates consists of three time-dependent signals, each for one spatial direction of the magnetic field, we decided not to use any feature that involves obvious correlations between the three signals. The reason is that when an array of sensors is utilized, all sensors axes must be aligned in order to be able to perform gradiometric analysis. Sensor axes' alignment can be performed by digital alignment of the 3 measured axes. The rotation matrix, which is different for each sensor in the array, is calculated by implementing a tilt-compensated electronic compass (also called "eCompass"). The full procedure and mathematics are described in [Ozyagcilar 2013]. However, this compensation results in mixing of the "pure" original fluxgate vector components data. Each axis signal is now a linear combination of the original signals. Therefore, although a dc jump should appear in only one axis at a time, due to the rotation matrix operation, it might be present now in one, two or even three axes at the same time. Therefore, we cannot consider the appearance of the dc jump in one axis at a time as a reasonable feature to employ. This feature, which could have helped us to discriminate dc jumps from another event that intrinsically involves more than one axis, becomes irrelevant.

## C. Training Process

The kernel-based method is optimized, and the width of the kernel is chosen as $\sigma = 2 \cdot F_s$, where $F_s$ is the sample frequency. This optimization is done via the training set, validated with the evaluation set and tested on the test set. The trade-off which facilitates this outcome is between distinguishing jumps from noise, which occurs as $\sigma$ decreases, and avoid mixing the statistical nature of noise in the decision rule, which may occur when $\sigma$ increases.

The learning-based models go through optimization that comprise two stages; the training process in which the parameters of the model are localized inside a narrow grid of values, and an evaluation stage which neat picks the best parameter set. While the SVM training process is done in a traditional manner, the ANN training process is worth describing in detail for future use of the reader. The network is initialized with weights drawn from a centered Gaussian distribution with variance of 0.01. It should be noticed that pre-training is not done in this study, and since the network is considered small, local minima is obviated with high probability. Optimization is employed by back-propagation through time, done by the classic gradient descent method. The parameters of the backpropagation are the learning rate of $10^{-2}$ and momentum of $0.9$. The cost function is proportional to the $\ell_2$ norm with regularization term, weighted by $\lambda = 10^{-2}$. The network was trained until either 150 epochs or minimum gradient value of $10^{-6}$ were achieved. The results given in this study always regard the objective test set, never seen before by the models.

## IV. RESULTS AND DISCUSSION

First, we investigate the difficulty of the kernel-based method to cope with cluttered environment. That is, we compare the performance of this method in both shielded and noisy, cluttered setups. Next, we wish to compare between the kernel-based and learning-based methods. Thus, we test all three approaches in real-world cluttered environment. The results of these two experiments are illustrated by the receiver operating curve (ROC). This curve allows us to examine a range of false positive rates, and their corresponding probabilities of detection. This trade-off is commonly used in detection systems, and quantitatively differs from the true negative versus false negative relation. Also, the SVM and ANN are tested head-to-head in two manners; first, we wish to deduce how well these methods handle low SNRs, which project on their robustness abilities. For that purpose, we decrease the value of SNR from 15 dB to 0 dB, while exploiting the entire training and evaluation sets. For each SNR value, we report the value that maximizes the summation of the true positive (TP) and true negative (TN) measures. This value is essentially the maximal accuracy rate. Second, we decrease the amount of training and evaluation sets size, while keeping their ratio unchanged and maintaining a level of 15 dB SNR. The target of this experiment is to inspect the generalization property of the learning-based methods. The same measure of TP+TN is extracted as in the previous experiment for each of the two approaches.

Additionally, we wish to examine the contribution of each of the 9 features to the performance of the algorithm. Here, we fix the conditions to 15 dB SNR and full amount of training and evaluation sets sizes. Then, for each of the two learning-based (and thus, feature-based) approaches, we perform the entire training, evaluation and test processes with increasing number of features, in a cumulative manner. Explicitly, we employ merely one feature, then two features, and so on, until we converge to the performance reported in the first experiment where we used all 9 features. It should be highlighted that for each number of features $1 \leq n \leq 9$, we performed $\binom{9}{n}$ experiments. For each experiment, the reported result is the one that generated the maximal accuracy. Again, the measure used here is TP + TN. By observing Fig. 2, we can conclude that the classic kernel-based method performs well when merely the internal noise of the sensor is of presence. However, when higher levels of noise and clutter are of presence, the ability of this method to cope with dc jumps detection is poor and cannot be considered sufficient for a reasonable application, which motivated the learning-based methods.

In continuation, we illustrate the advantages of the learning-based methods in Fig. 2. Initially, one can observe an enhanced performance of the latter in comparison to the kernel-based method. This improvement projects on both the high generalization and the robustness abilities that the learning-based methods possess, and the kernel-based method lacks. By focusing on the learning-based methods, we can spot the rapid convergence of the ANN against the SVM. Namely, the ANN produces a better separating model, which was expected due to the ability of deep-learning methods to model highly complex non-linear relations of data, while the SVM is restricted to shallower and more limited non-linear patterns.

To compare the two learning-based methods in a more profound way, we observe the results of two experiments, demonstrated in the top and middle graphs in Fig. 3. In the former, we notice that the lower the SNR, the worse the SVM method can cope and perform. This indicates on the enhanced robustness that the ANN brings, which can be affiliated with the large number of perceptron units and the contribution of each of them for the end-to-end non-linear model of the network. In the latter, the power of the ANN is again revealed, now regarding the desired generalization property, i.e., the ability to perform well on unseen data efficiently. Due to the dropout and regularization techniques we applied during training to yield the ANN model, the latter can extract the intrinsic properties of the training data instead of wasting coefficients and model the noise. Therefore, while low amount of training data causes the SVM to learn the trend of the noise, the ANN is able to neglect it and lower the influence of noise on the dc jumps detection performance.

Eventually, an interesting conclusion can be deduced based on the bottom graph in Fig. 3. The x axis ticks label $L_i$, $1 \leq i \leq 9$, represents the subset of features from $f_1$ to $f_i$, that are used for training and evaluation. First, only four out of nine features are enough to yield 90% accuracy for the ANN; correlation with step function, correlation with the dedicated kernel $K(n,\sigma)$, correlation to previous time frames and kurtosis. However, to reach this accuracy, the SVM exploits all nine features. Also, both learning-based methods experience similar rate of increase in accuracy regarding the cumulative feature analysis. This is expected, since these methods essentially attempt to minimize a similar optimization problem.

## V. CONCLUSION

In this work we have performed dc jumps detection with three methods, fed by physical, statistical and temporal features. We have shown that the classic kernel-based approach cannot comprehend non-shielded environment and demonstrated the enhanced abilities of the learning-based methods in detecting dc jumps when high levels of noise and clutter are present. We performed dedicated experiments to analyze the robustness and generalization properties of the SVM and ANN based approaches. We deduced that the deep ANN architecture is able to cope with low levels of SNR, with relatively small amount of data seen during training. Conclusively, the ANN-based detection system has low sensitivity to the low signal fidelity and high levels of inherent noise caused by the Barkhausen phenomenon. Future work will involve implementation of the suggested detection methods in an online platform that employs magnetic sensing.

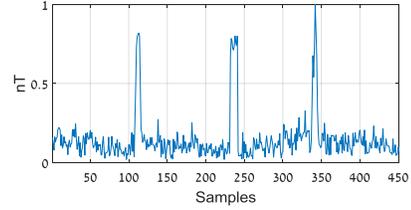

Fig. 1. Magnetic noise in low power parallel fluxgate magnetometer with a dc jump (samples 230-240), and two other clutter-based anomalies that bear high resemblance to dc jump.

Table 1. Features Description.

| NOTATION | RANGE | DESCRIPTION |
|---|---|---|
| $f_1$ | [0 1] | Correlation with step function |
| $f_2$ | [0 1] | Correlation with $k(n,\sigma)$ |
| $f_3$ | [0 1] | Correlation of adjacent frames |
| $f_4$ | [0 ∞] | Kurtosis |
| $f_5$ | [0 1] | Pearson's coeff. [Lawrence 1989] |
| $f_6$ | [0 ∞] | Skewness |
| $f_7$ | [0 1] | Covariance |
| $f_8$ | [0 1] | Carmer's V [Wu 2013] |
| $f_9$ | [0 1] | Spearman's coeff. [Zar 1972] |

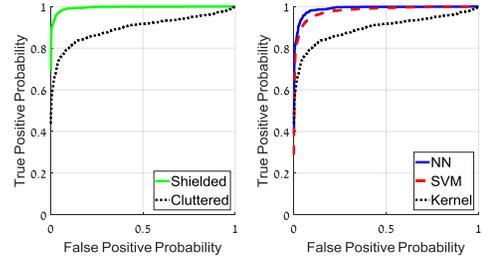

Fig. 2. ROC of kernel method in shielded and cluttered environment (left), and the three detection methods in cluttered environment.

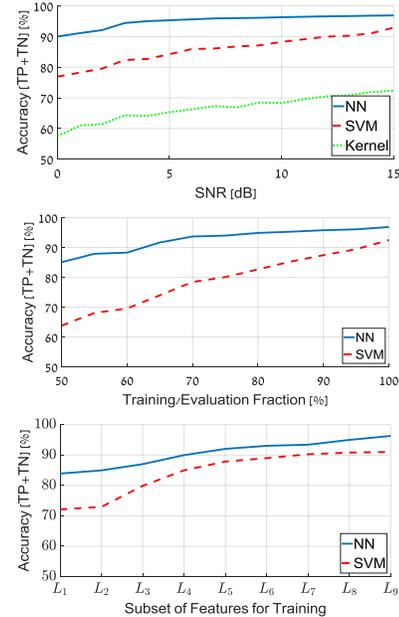

Fig. 3. Accuracies (TP+TN) of the detection methods versus different values of SNR (top), fractions of full-sized training set used for training (middle), and subsets of features used for training (bottom).